\documentclass[final,5p,times,twocolumn,numbers]{elsarticle}

\usepackage{amssymb}
\usepackage{amsmath}
\usepackage{booktabs}
\usepackage[ruled,vlined]{algorithm2e}
\usepackage{multirow}
\usepackage{colortbl} 
\usepackage{graphicx}
\usepackage[table]{xcolor} 

\usepackage{lineno}

\usepackage{hyperref}
\hypersetup{colorlinks=true, linkcolor=blue, anchorcolor=blue, citecolor=blue}
\usepackage{cleveref}

\usepackage{array, caption}
\usepackage[font=small,labelfont=bf,labelsep=none]{caption}
\journal{Pattern Recognition}
\begin{document}

\begin{frontmatter}


\title{Towards Unbiased Source-Free Object Detection via  Vision Foundation Models} 

\author[a,b]{Zhi Cai}
\ead{caizhi97@buaa.edu.cn}

\author[a,b]{Yingjie Gao}
\ead{gaoyingjie@buaa.edu.cn}

\author[c]{Yanan Zhang}
\ead{yananzhang@hfut.edu.cn}

\author[a,b]{Xinzhu Ma}
\ead{xinzhuma@buaa.edu.cn}

\author[a,b]{Di Huang\corref{correspond}}
\ead{dhuang@buaa.edu.cn}

\cortext[correspond]{Corresponding author.}

\affiliation[a]{State Key Laboratory of Complex and Critical Software Environment, Beihang University, Beijing, 100191, China}
\affiliation[b]{School of Computer Science and Engineering, Beihang University, Beijing, 100191, China}
\affiliation[c]{School of Computer Science and Information Engineering, Hefei University of Technology, Hefei, 230601, China}

\begin{abstract}

Source-Free Object Detection (SFOD)  has garnered much attention in recent years  by eliminating the need of source-domain data in cross-domain tasks, but existing SFOD methods suffer from the Source Bias problem, \emph{i.e.} the adapted model remains skewed towards the source domain, leading to poor generalization and error accumulation during self-training.
To overcome this challenge, we propose Debiased Source-free Object Detection (DSOD), a novel VFM-assisted SFOD framework that can effectively mitigate source bias with the help of powerful VFMs. Specifically, we propose Unified Feature Injection (UFI) module that integrates VFM features into the CNN backbone through Simple-Scale Extension (SSE) and Domain-aware Adaptive Weighting (DAAW). Then, we propose Semantic-aware Feature Regularization (SAFR) that constrains feature learning to prevent overfitting to source domain characteristics.
Furthermore, we propose a VFM-free variant, termed DSOD-distill for computation-restricted scenarios through a novel Dual-Teacher distillation scheme.
Extensive experiments on multiple benchmarks demonstrate that DSOD  outperforms state-of-the-art SFOD methods, achieving 48.1\% AP on Normal-to-Foggy weather adaptation, 39.3\% AP on Cross-scene adaptation, and 61.4\% AP on Synthetic-to-Real adaptation.
\end{abstract}




\end{frontmatter}




\section{Introduction}



Unsupervised Domain Adaptive Object Detection (UDAOD)~\cite{swfaster,aqt,mttrans,dadetr} has attracted significant research attention in recent years, primarily driven by the critical challenge of domain shift in real-life applications. The domain shift problem arises when there is a distribution discrepancy between the training (source domain) and testing (target domain) data, leading to substantial performance degradation.  However, a fundamental limitation of UDAOD approaches is their dependence on source domain data during adaptation, which introduces additional practical constraints regarding data storage, privacy, and computational requirements.
In contrast, Source-Free Object Detection (SFOD)~\cite{sed,irg,a2sfod,dru,lpld} represents a more practical and challenging paradigm that eliminates the need for source data during adaptation.

\begin{figure}[t]
    \centering
    \includegraphics[width=1.0\linewidth]{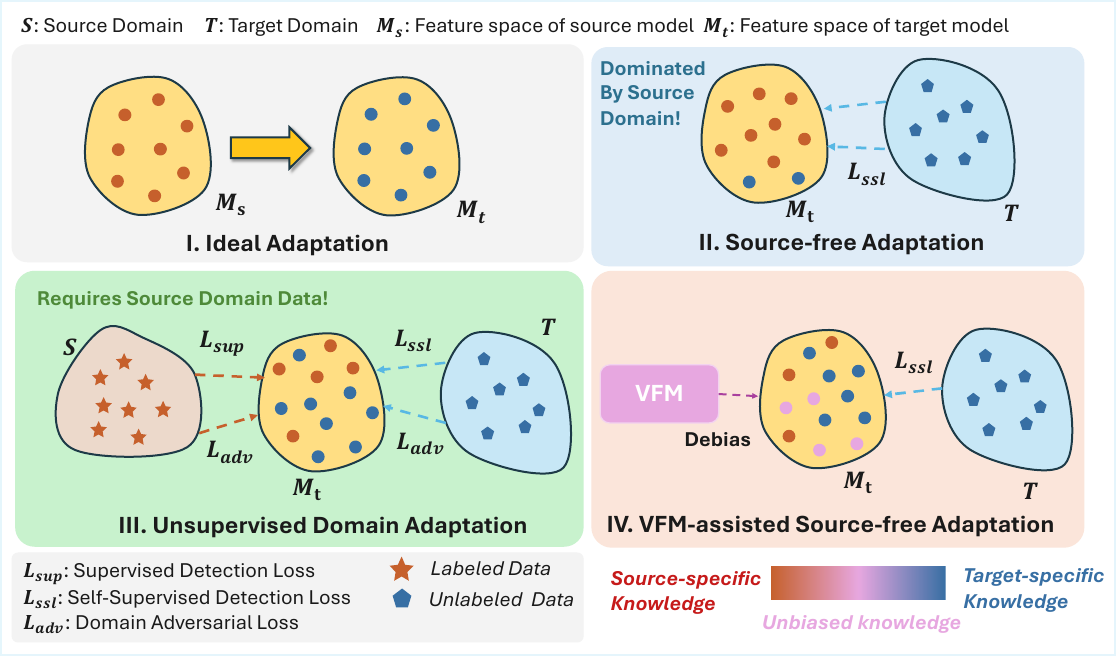}
    \caption{Conceptual comparison of different domain adaptation paradigms.
(I) Ideal Adaptation: Achieves complete knowledge transfer from the source domain (\textcolor{red}{red}) to the target domain (\textcolor{blue}{blue}).
(II) Traditional SFOD: Prone to source bias, as indicated by the predominance of red circles representing source-domain characteristics.
(III) UDAOD: Enables more effective adaptation but requires access to source-domain data.
(IV) VFM-assisted SFOD: Mitigates source bias by incorporating unbiased knowledge (\textcolor{purple}{purple}) while eliminating the need for source-domain data access.}
    \label{fig:fig1}
\end{figure}

Specifically, SFOD methods~\cite{irg,a2sfod,dru,lpld} operate under the constraint that only a source-pretrained model and unlabeled target data are available. Compared to UDAOD methods~\cite{swfaster,aqt,mttrans,dadetr}, SFOD approaches are more prone to the \textbf{Source Bias} problem (model are biased towards source rather than target domain), as illustrated in Fig.~\ref{fig:fig1} (I,II,III). One key reason is that UDAOD methods can learn domain-invariant features~\cite{tia} by leveraging both source and target domain data during training, which facilitates effective knowledge transfer from the source domain to the target domain. In contrast, SFOD relies solely on unlabeled target data, which restricts the model's ability to capture robust cross-domain representations and leads to suboptimal knowledge transfer. Consequently, SFOD models often exhibit poorer generalization performance and are more susceptible to error accumulation throughout the self-training process.




The recent emergence of Vision Foundation Models (VFMs), such as DINOv2~\citep{oquab2023dinov2}, offers a promising direction to address the mentioned challenges. We argue that integrating VFMs into SFOD provides two key advantages: \textbf{Semantic Complementarity} and \textbf{Regularization Effect}. First, VFMs offer generic and robust visual representations, while source-pretrained models provide instance-level and class-aware features, enabling complementary representation learning. Second, the frozen VFM can serve as a regularizer, preventing the model from overfitting to narrow source-domain patterns. By leveraging more diverse and unbiased information, this approach mitigates source bias and enhances generalization performance.



In this paper, we propose a novel method named \textbf{Debiased Source-Free Object Detection (DSOD)} that mitigates  source bias by leveraging the capability of VFMs, as depicted in Fig.~\ref{fig:fig1} (IV). It adopts a dual-tower architecture that comprises a DINOv2~\cite{oquab2023dinov2} encoder and a standard ResNet~\cite{he2016deep} encoder to achieve less biased adaptation.
To enrich the output features with unbiased knowledge,  we introduce an \textbf{{U}nbiased {F}eature {I}njection (UFI)} mechanism, which enhances the semantic richness of original CNN features by integrating features extracted from DINOv2. Then we propose a \textbf{Semantic-Aware Feature Regularization (SAFR)} module to strengthen the intrinsic correlation between the CNN and VFM encoders, thereby further eliminating feature-level bias. Besides, to alleviate training instability, we design a \textbf{Domain-Aware Adaptive Weighting (DAAW)} strategy that adaptively determines fusion weights via a heuristic approach and employs gradual weighting to smooth optimization conflicts.
To facilitate the deployment in the edge devices, we further introduce a \textbf{Dual-Teacher (DT)} distillation scheme. It helps distill the knowledge from DSOD to a VFM-free variant, named DSOD-Distill, that incurs \textbf{no extra inference burden}. 

To summarize, our contributions are five-fold:
\begin{enumerate}
    \item We present DSOD, a novel framework that significantly enhances SFOD performance by exploiting VFMs for complementary semantics and regularization.
    \item We design two core modules, UFI and SAFR. The former injects semantic-rich features of VFM into the detection model, and the latter restricts the manifold learning of CNN encoders by semantic-aware regularization from VFMs.
    \item  We further introduce a DAAW strategy to automate fusion weight selection and stabilize the  training with smooth weighting schedule.
    \item We devise a DT method to transfer the VFM's knowledge into a compact, VFM-free model (DSOD-Distill), enabling practical deployment.
    \item Extensive experiments on multiple benchmarks demonstrate the effectiveness of the proposed method.
\end{enumerate}

\section{Related Works}
\subsection{Object Detection}
The remarkable success of deep neural networks in image classification~\cite{he2016deep} has spurred the development of CNN-based detectors, which can be broadly categorized into {\emph{two-stage}~\cite{faster}, \emph{one-stage}~\cite{focalloss} and \emph{anchor-free} methods~\cite{tian2020fcos, zhou2019objects}.}
As a representative two-stage approach, Faster R-CNN~\citep{faster} employs a 
\emph{Region Proposal Network} (RPN) to generate region proposals, which are subsequently refined by an ROI head.
While effective, this framework suffers from inherent limitations in efficiency and generalization due to its multi-stage design and reliance on hand-crafted anchor boxes.
In contrast, single-stage methods~\cite{liu2016ssd,focalloss} integrate proposal generation and detection into a single forward pass, achieving superior computational efficiency and simplicity.
Anchor-free methods advance this direction further by predicting objects directly from key points such as the center~\cite{tian2020fcos} or corners~\cite{law2018cornernet}.
Despite these advancements, most CNN-based detectors still produce multiple predictions per object and rely on the non-maximum suppression (NMS) strategy to remove redundant detections.

More recently, the proposal of DETR~\cite{detr} reformulates the pipeline of detection to a set-prediction problem with the introduce of transformer~\cite{vaswani2017}. 
Designed with an end-to-end paradigm, DETR totally eliminates the need of hand-crafted components such as NMS and anchor boxes.
However, it suffers from slow-convergence issue.  Following work~\cite{deformabledetr} explores utilizing sparse attention in transformer layer and achieves impressive progress. 
Some works~\cite{dino,hdetr} suppose the slow-convergence issue is also due to the one-to-one matching and improve DETR with query-denoising strategy~\cite{ dino} or many-to-one matching~\cite{hdetr}.  
Some works also focus on improving the loss design by introducing IoU-aware loss items~\cite{cai2024align}.

In this work, we select Deformable-DETR~\cite{deformabledetr} as our baseline for its strong performance and fast convergence. 



\subsection{Domain Adaptive Object Detection}
Current UDAOD methods roughly fall in two categories: feature alignment methods~\cite{swfaster,tia,dafaster,robust_object_detection} and self-training methods~\cite{mttrans, mtm}. The former seek to align the image features of target domain to those in source domain and thus learn a domain-agnostic feature space~\cite{swfaster,dafaster}. The latter focus on progressively improving models by generating pseudo labels on target domain~\cite{mttrans, mtm}. These works have made huge advancement in the UDAOD setting. However, the access to source domain data  may violates data privacy in some real-word scenarios. 

Source-free object detection builds upon the idea of UDAOD while extending its framework to address data privacy concerns. As the name implies, SFOD operates without access to source domain data. 
To overcome such a challenge, most works~\cite{sed,a2sfod,pets,dru} adopt a self-training paradigm such as the MT framework~\cite{meanteacher}.
SED~\cite{sed} is the first method that works on the SFOD problem, which proposes an entropy-based threshold searching mechanism to filter fairly reliable pseudo-labels from noisy ones. 
Subsequently, A2SFOD~\cite{a2sfod} proposes splitting the target data into source-similar and source-dissimilar subsets using a variance-based criterion, and then applies a domain classifier to learn domain-agnostic features.
After that, IRG~\cite{irg} introduces an instance relation graph network equipped with contrastive loss to enhance the feature representation. 
PETS~\cite{pets} introduces a multiple-teacher framework that consists of a dynamic teacher and a static teacher.
Some works~\cite{lpu,lpld} exploit the utilization of low-quality pseudo labels.
Recently, DRU~\cite{dru} proposes a dynamic retraining strategy to alleviate the performance degradation issue commonly encountered in MT frameworks. DDT~\cite{ddt} argues different weights should update differently and propose an asymmetric weight updating strategy to replace symmetric operation.
Despite these advancements, the SFOD methods fall behind UDAOD methods in learning domain-agnostic feature representation due to the absence of source domain data, which leads to severe source bias issue.
In this work,  we leverage the  powerful and representative features of VFM  to compensate the absence of source data and mitigate the source bias issue.

\subsection{Vision Foundation Model-Assisted Object Detection}

Vision foundation models refer to a class of models designed for fundamental vision tasks, such as generalizable feature extraction~\cite{oquab2023dinov2} and reference segmentation~\cite{ma2024segment}. Due to their rich and transferable representations, VFMs have become ubiquitous in object detection research~\cite{fu2024frozen,crowdsam}.
Current approaches to integrating VFMs into detection frameworks can be broadly categorized into two types. One line of work, exemplified by Frozen-DETR~\cite{fu2024frozen}, employs a pre-trained VFM like DINOv2~\cite{dino} as an auxiliary backbone to enhance the visual understanding of DETR-based detectors. Another line of methods~\cite{vfmadapter,vitadapterdp} adopts VFMs as the primary backbone and introduces carefully designed adapters to improve their dense prediction capabilities. For instance, ViT-Adapter~\cite{vitadapterdp} proposes a dual-encoder architecture combining a CNN and a ViT~\cite{dosovitskiy2020image} to adapt vision transformers for dense tasks.
In the context of UDAOD, DINO-Teacher~\cite{dinoteacher} leverages DINOv2 as an offline pseudo-labeler on the target domain and aligns CNN features with VFM representations to enhance domain adaptation.
Despite these advances, no previous work has explored integrating a VFM backbone into an existing source-pre-trained detection model. In this paper, we demonstrate through extensive experiments that this integration represents a highly promising research direction.


\section{Preliminary}
\subsection{Problem Formulation}

We first present the problem definition of SFOD. 
Let  $D_s=\{X_s, Y_s\}$ denote source domain data and $D_t=\{X_t\}$ denote target domain data. 
Here we use $X_s$ and $X_t$ to denote the images, and $Y_s $  to denote the annotations. 
The goal of SFOD is to adapt  a detection model  $G$ pre-trained on $D_s$ to unlabeled target domain data $D_t$.
The core challenges of SFOD are twofold: (1) the absence of labels in the target domain, and (2) the inaccessibility of the source domain data during adaptation.
\subsection{Detection Transformer}\label{sec:detr}
DETR~\cite{detr} reformulates the problem of object detection into a set prediction problem and thus is able to remove the hand-crafted process such as Non-maximum Suppression (NMS) and anchors. The DETR series~\cite{detr,deformabledetr,dino,cai2024align} follow a classical framework that consists of a CNN encoder, a transformer encoder and a transformer decoder. The optimization goal for DETR is given as a set matching form:
\begin{equation}
    \label{eq:loss_detr}
     \mathcal{L}_{\text{det}} = \sum_i \Bigl[ \alpha \cdot \mathcal{L}_{\text{cls}}(c_{\sigma(i)} , \hat{c}_{i}) + \beta \cdot \mathcal{L}_{\text{bbox}}(b_{\sigma(i)},\hat{b}_{i}) \Bigl],
\end{equation}
where $(c,b)$ and $(\hat{c},\hat{b})$ represent the ground truth and predictions, respectively; $\alpha,\beta$ are two weighting coefficients; $\mathcal{L}_{\text{cls}}$ is a classification loss (\emph{e.g.}, focal loss~\cite{focalloss}), $\mathcal{L}_{\text{bbox}}$ is a combination of L1 loss and GIoU loss; and $\sigma(i)$ represents the optimal matching ground truth index for the $i$-th prediction. 
\subsection{Mean-Teacher Framework}\label{sec:mt}
The Mean-Teacher framework~\cite{meanteacher} was first introduced  for Semi-Supervised Learning (SSL) and has since been widely adopted in SFOD~\cite{irg,a2sfod,lpld,dru,ddt} for its impressive performance.
This framework consists of a student model $G_{\hat{\theta}}^{stu}$ and a teacher model $G_{\theta}^{tch}$ which share the same network structure. Both two models are initialized with identical weights. During training, the model works in a self-improving manner: the student learns from unlabeled data using  pseudo labels provided by the teacher, while the weights of the teacher are updated by via Exponential Moving Average (EMA) of the student's weights:
\begin{equation}
    \Theta_{tch}^{t+1} =  \alpha \cdot \Theta_{tch}^t  +  (1- \alpha) \cdot \Theta_{stu}^t ,
\end{equation}
where $\alpha$ is a momentum parameter that controls the speed of update and $t$ is the time step. Also, a confidence hyper-parameter $\delta$ is adopted for filtering low-confident samples. To enhance the stability and robustness, the teacher network receives the weakly augmented images to produce reliable pseudo-labels, whereas the student is fed with strongly augmented images to encourage  invariant and discriminative features.

\begin{figure}[t]
    \centering
    \includegraphics[width=\linewidth]{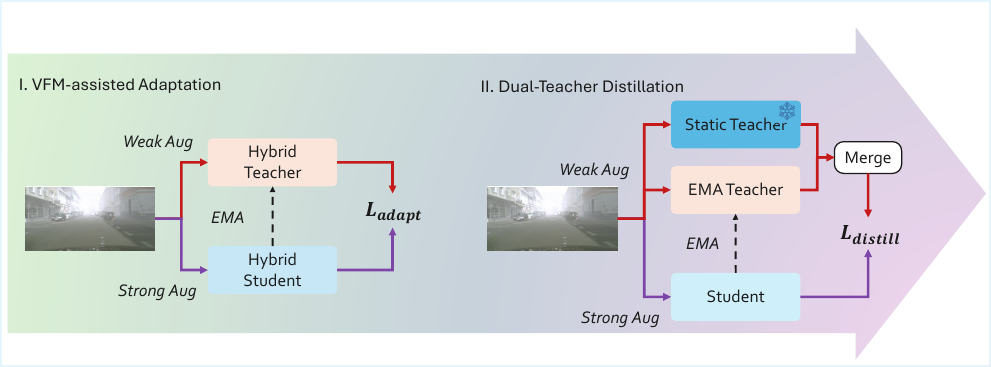}
    \caption{Overview of our method DSOD and DSOD-Distill. In the  Stage I, we train a hybrid model (DSOD) on target-domain data with the assistance of VFM. Then, in Stage II, we transfer the knowledge from DSOD to a VFM-free variant, DSOD-Distill through Dual-Teacher distillation.}
    \label{fig:stage_distill}
\end{figure}

\begin{figure*}[!t]
    \centering
    \includegraphics[width=1.0\linewidth]{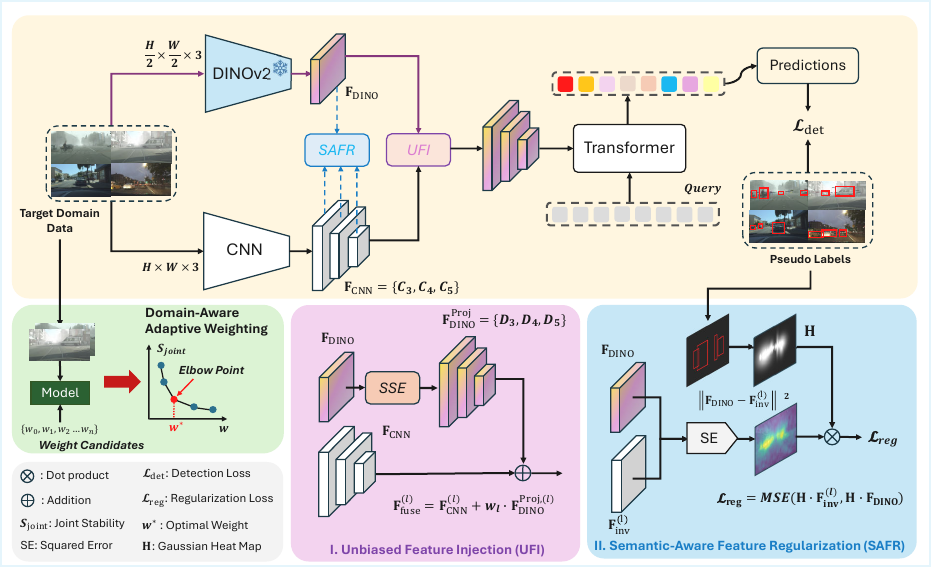}
    \caption{The main components (student-part) of DSOD.  To utilize the generalizable and semantic-rich features, we operate in two aspect: (I) Unbiased Feature Injection (UFI) and (II) Semantic-Aware Feature Regularization (SAFR). Also, we propose a Domain-Aware Adaptive Weighting (DAAW) strategy to select proper weighting and stablize the training. }
    \label{fig:fig_main}
\end{figure*}

\section{The Proposed Method}
As shown in Fig. \ref{fig:stage_distill} (I), DSOD is built upon DETR and the Mean Teacher (MT) framework. Leveraging this paradigm, we train a VFM enhanced model (hybrid model) for the SFOD task. Fig. \ref{fig:fig_main} illustrates the details of the hybrid model. Specifically, we utilize DINOv2~\cite{oquab2023dinov2} as the auxiliary encoder for its rich semantic features. Our method functions on two complementary levels: (1) the UFI module performs feature fusion by injecting VFM features to mitigate source bias (Sec.~\ref{sec:ufi}), and (2) the SAFR module provides explicit regularization to enhance the CNN backbone's representations (Sec.~\ref{sec:safr}). Training instability is addressed by the DAAW strategy (Sec.~\ref{sec:daaw}). For practical application, the DT distillation scheme is proposed for resource-constrained devices (Fig. \ref{fig:stage_distill} (II) and Sec.~\ref{sec:dt}).


\subsection{Unbiased Feature Injection }\label{sec:ufi}

A obstacle to integrating the VFM with the source-pretrained object detector lies in their \textbf{architectural mismatch}. For instance, the CNN backbone is hierarchical while  the ViT backbone, which is widely adopted by VFMs,  is plain.  
To address this discrepancy,  we introduce a lightweight  \textbf{Simple Scale Extension (SSE)}  module that aligns DINOv2 features with CNN features in both channel and spatial dimensions. 

The design of  UFI is depicted in Fig.~\ref{fig:fig_main} (I). For the given input image $X \in R^{ H \times W \times 3}$, the CNN backbone produces multi-level features denoted as $\mathbf{F}_{\text{CNN}} =\{C_{3} ,C_{4},C_{5}\}$ in the stride of \{8,16,32\}. The DINOv2 backbone is plain and predict a single-level feature map denoted as $\mathbf{F}_{\text{DINO}}$. We use several linear projection layers to project it to the CNN dimension for each scale $l$:
\begin{equation}
    \mathbf{F}_{\text{DINO}}^{ \text{Proj},(l)} = f(\text{Proj}^{(l)}(\mathbf{F}_{\text{DINO}})),
\end{equation}
where $\text{Proj}^{(l)}$ implies a linear transformation implemented via a 1 $\times$ 1 convolution and $f$ represents a bilinear interpolation. Through this process, we align both the spatial and channel dimension of $\mathbf{F}_{\text{DINO}}$ with the target feature $\mathbf{F}_{\text{CNN}}$.
\begin{figure}[ht]
    \centering
    \includegraphics[width=0.75\linewidth]{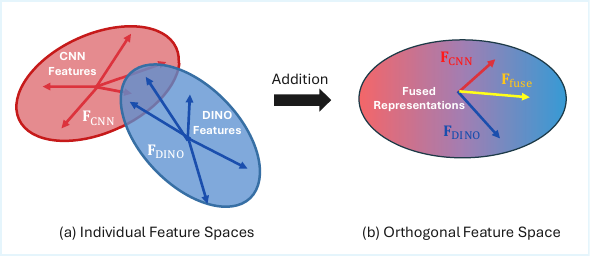}
    \caption{Illustration of the orthogonal feature space. The fused feature vector \textcolor{orange}{$\textbf{F}_{\text{fuse}}$}, obtained through addition, preserves the information from both original \textcolor{red}{$\textbf{F}_{\text{CNN}}$} and  \textcolor{blue}{$\textbf{F}_{\text{DINO}}$}.}
    \label{fig:idea_of_orthogonal}
\end{figure}

The aligned features are fused via a simple yet effective additive scheme:
\begin{equation}
\label{eq:fusion}
    \mathbf{F}_{\text{fuse}}^{(l)} = \mathbf{F}_{\text{CNN}}^{(l)} + \mathbf{F}_{\text{DINO}}^{\text{Proj},(l)} \cdot w_l,
\end{equation}
where $w_l$ is a layer-wise weighting factor. We also design a self-adaptive adjustment mechanism for this factor to stabilize the training (see Sec.~\ref{sec:daaw}). Note that we use addition to achieve feature fusion, while architecturally simple, we argue that this design have following advantages:
\textbf{First}, we leverage the property of high-dimensional feature orthogonality: vectors from independently pre-trained models are approximately orthogonal~\cite{vershynin2018high}, allowing additive fusion to preserve the principal directions of the original ResNet~\cite{he2016deep} features while seamlessly integrating complementary DINOv2~\cite{oquab2023dinov2} semantics (Fig.~\ref{fig:idea_of_orthogonal}). \textbf{Second}, this minimalist design introduces negligible parameters, significantly reducing overfitting risks in data-scarce target domains.
We provide extensive ablation studies (Sec.~\ref{sec:ablation}) and visual analysis (Sec.~\ref{sec:visual_analysis}) to support our claims. 

\subsection{Semantic-Aware Feature Regularization}\label{sec:safr}
Beyond leveraging VFM features for semantic complementarity, we further enhance feature robustness through foreground-aware regularization. Unlike prior works~\cite{dinoteacher} that treat all spatial positions equally, our method explicitly prioritizes foreground regions where class-aware information is most salient. By strengthening the feature representations in these discriminative areas, the category decision boundaries become more tightly constrained by the less-biased VFM features, resulting in a more robust and debiased encoder.


Specifically, we inversely align the CNN features with the DINOv2 features by projecting and resizing them via an MLP and bilinear interpolation:
\begin{equation}
    \mathbf{F}_{\text{inv}}^{(l)} = f(\text{Proj}^{(l)}(\mathbf{F}_{\text{CNN}}^{(l)})),
\end{equation}
where $\mathbf{F}^{(l)}_{\text{inv}}$ are the inversely projected features, $\text{Proj}^{(l)}$ implies a two-layer MLP block and $f$ denotes the bilinear interpolation function.
Then, we access the pseudo-labels generated by teacher (See Sec.~\ref{sec:mt}) for salient regions. The pseudo labels are then converted into a class-agnostic heatmap. This conversion is performed using a Gaussian distribution:
\begin{equation}
\mathbf{H}^{(i,j)} = \max_{b \in \mathcal{B}} \exp\left(-\frac{1}{2} \left[ \frac{(i - b_x)^2}{{b_w}^2} + \frac{(j - b_y)^2}{{b_h}^2} \right]\right),
\end{equation}
where $b_x$, $b_y$, $b_w$, $b_h$ represent the center point coordinate, width and height of a bounding box, \(\mathcal{B}\) denotes the set of all pseudo-label bounding boxes, and \((i, j)\) are spatial coordinates in the feature map.  The $\max$ operator selects the maximum response across all bounding boxes, thereby prioritizing the most salient foreground region at each spatial location. We visualize this process in Fig.~\ref{fig:fig_main} (II).
  
Finally, we can define a weighted Mean Squared Error (MSE) loss as the regularization loss:
\begin{equation}
    \mathcal{L}_{\text{reg}} = \frac{1}{HWL} \sum_{i=1}^{H} \sum_{j=1}^{W} \sum_{l=1}^{L} \mathbf{H}^{(i,j)} \cdot || (\mathbf{F}_{\text{inv}}^{(i,j,l)} - \mathbf{F}_{\text{DINO}}^{(i,j)}||^2,
\end{equation}
where $H$,$W$ denote the height and width of DINO's feature map, $L$ represents the number of feature maps, $(i,j)$ are spatial coordinates, and $l$ indexes the feature map.



\subsection{Domain-Aware Adaptive Weighting}\label{sec:daaw}

\begin{figure}[ht]
    \centering
    \includegraphics[width=\linewidth]{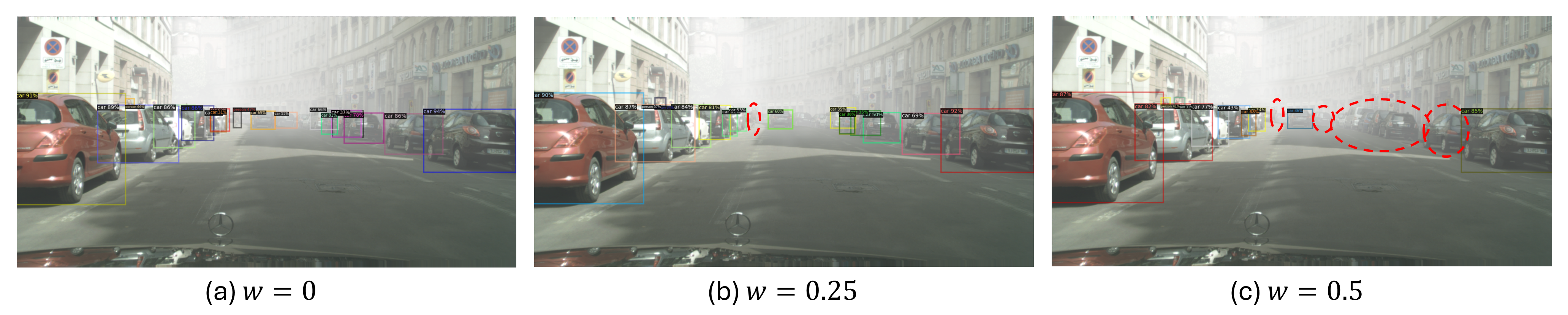}
    \caption{Visualization of prediction instability under different fusion weights. \textcolor{red}{Red circle} highlights the regions where objects are missing compared to the non-fusion baseline model ($w$=0).}
    \label{fig:unstable_pred}
\end{figure}

During the initial adaptation process, the  fused model is prone to training instability, which refers to the issue that fused model produces low-quality predictions as illustrated in Fig.~\ref{fig:unstable_pred}. 
This stems from the fact that the merged feature manifold deviates from the original, leading the model to produce unpredictable outputs. This instability is further amplified by the self-training nature of the MT framework, which can propagate and reinforce these initial errors throughout the adaptation process.

To address this issue, we propose to conduct a sensitivity test and select elbow point (with maximum curvature) as the fusion weights,  as depicted in Fig.~\ref{fig:fig1} (\textcolor{green}{green part}).  Given $N$ randomly selected target domain test samples $\mathcal{D}_{\text{tgt}} = \{I_0, I_1,I_2,...,I_n\}$, we compare the prediction consistency between the initial source-pretrained model $\mathcal{M}_{\text{init}}$ and the fused model $\mathcal{M}_{\text{fuse}}$. We first select the $K$ most confident instances from $\mathcal{M}_{\text{init}}$ and use them for pairwise evaluation.   Specifically, we define the following three metrics.


\noindent\textbf{Classification Stability } measures the consistency of class predictions:
\begin{equation}
\mathcal{S}_{\text{cls}} = 1 - \frac{1}{K} \sum_{k=1}^{K} \left| \frac{{c}_k^{\text{init}} - {c}_k^{\text{fuse}}}{{c}_k^{\text{init}}} \right|,
\label{eq:cls_stab}
\end{equation}
where $ c_k^{\text{init}} $ and $ c_k^{\text{fuse}}$ denote the confidence scores for the predicted class $c_k$ of the $k$-th selected instance from the initial and new models, respectively. $\mathcal{S}_{\text{cls}}$ is normalized by the initial score $c_k^{\text{init}}$ to balance the influence of both high-confident samples and low-confident samples.

\noindent\textbf{Localization Stability } measures the consistency of bounding box predictions:
\begin{equation}
\mathcal{S}_{\text{loc}} = \frac{1}{K} \sum_{k=1}^{K} \text{IoU}\left(\mathbf{b}^{\text{init}}_{k}, \mathbf{b}^{\text{fuse}}_{k}\right),
\label{eq:loc_stab}
\end{equation}
where $\text{IoU}(\cdot, \cdot)$ computes the Intersection over Union (IoU) between two bounding boxes, and $\mathbf{b}_{\text{init}}^{(k)}, \mathbf{b}_{\text{new}}^{(k)}$ are the corresponding bounding boxes of the $k$-th instance.

\noindent\textbf{Joint Stability} measures the overall consistency by combining both classification and localization stability:
\begin{equation}
\mathcal{S}_{\text{joint}} = \sqrt{\mathcal{S}_{\text{cls}} \cdot \mathcal{S}_{\text{loc}}},
\label{eq:joint_stab}
\end{equation}
where  geometric mean is chosen over the arithmetic mean because it more severely penalizes significant drops in either component.
The complete selection procedure of $w_l^*$ is outlined in Algorithm~\ref{alg:weight_select}.  We conduct sensitivity test first and then apply  the elbow-criterion to  find the elbow point. This elbow point represents the transition where additional VFM feature incorporation yields diminishing returns relative to the introduced instability. We validate the effectiveness of such process with experimental results (Sec.~\ref{sec:ablation}).

\begin{algorithm}[!t]
\caption{Weight Selection via Stability Evaluation}
\label{alg:weight_select}
\SetAlgoLined
\KwIn{Model $M_{\text{init}}$: pre-trained source model \\
      Model $M_{\text{fuse}}(w)$: fused model \\
      Image set $\mathcal{D}_{\text{tgt}} = \{I_1, I_2, \dots, I_N\}$ \\
      Weight candidates $\mathcal{W} = \{w_1, w_2, \dots, w_T\}$ }
\KwOut{Selected fusion weight $w^*$}

$\mathbf{S} \gets [\;]$ \tcp*{Initialize stability scores list}

\ForEach{$t$ in $range(1,T)$}{
  $\text{total\_score} \gets 0$ \\
  \For{$i \gets 1$ \KwTo $N$}{
    $\mathbf{P}_{\text{init}}^{(i)} \gets M_{\text{init}}(I_i)$ \tcp*{Run initial model}
    $\mathbf{P}_{\text{fuse}}^{(i)} \gets M_{\text{fuse}}(w_k)(I_i)$ \tcp*{Run fusion model with weight $w$}
    $\mathcal{S}_{\text{joint}}^{(i)} \gets \text{Stability}(\mathbf{P}_{\text{init}}^{(i)}, \mathbf{P}_{\text{fuse}}^{(i)})$ \tcp*{See Eq.~\ref{eq:cls_stab},~\ref{eq:loc_stab},~\ref{eq:joint_stab}}
    $\text{total\_score} \gets \text{total\_score} + \mathcal{S}_{\text{joint}}^{(i)}$ 
  }
  $\mathbf{S}[t] \gets \text{total\_score} / N$ \tcp*{Average over images}
}

$\Delta S \gets \text{Diff}(\mathbf{S}, 2)$ \tcp*{Compute second-order finite differences}

$t^* \gets \underset{t}{\text{argmax}}\ \Delta S[t]$ \tcp*{Find index of maximum curvature}

$w^* \gets \mathcal{W}[t^*]$ \tcp*{Select corresponding weight}

\Return{$w^*$}
\end{algorithm}

After the selection of $w_l^*$,  we employ a square root warm‑up weighting schedule for stable training:
\begin{equation}
w_l = \min (\sqrt{\tfrac{i}{N_\text{warm}}} ,1) \cdot w_l^*
\label{eq:weight_warmup},
\end{equation}
where $i$ denotes the current training iteration, $N_{\text{warm}}$ is the number of iterations for warm-up, which is empirically set to the length of one epoch, and $w_l^*$ is the target fusion weight obtained via sensitivity evaluation. This scheme smoothly increases the influence of DINO features from zero to the full target weight $w_l^*$, mitigating abrupt changes in the feature manifold and improving training stability.





\subsection{Dual-Teacher Distillation Scheme} \label{sec:dt}
Since the dual-tower architecture introduces additional computational overhead during both training and inference, it may be unsuitable for scenarios with limited computational resources. As illustrated in Fig.~\ref{fig:stage_distill}, we propose an optional two-stage scheme, in which the knowledge‑fused dual‑tower model serves as an additional frozen teacher to assist a conventional model during self‑training. This design retains the benefits of incorporating VFM‑based knowledge while avoiding extra computational costs. We refer to this variant as \textbf{DSOD‑Distill}.

During the distillation stage, we adopt a dual-teacher architecture,  inspired by PETS~\cite{pets}, utilizing both an EMA teacher (VFM‑free) and the DSOD teacher (the adapted DSOD model), and fuse their supervision signals. This design enables the model to benefit from the continuously updated supervision of the EMA teacher while also receiving efficient and unbiased supervision from the DSOD model. 
Unlike prior works~\cite{pets} that rely on box-level fusion, we introduce a loss-level fusion strategy. This approach more effectively preserves the complementary information from both teachers by operating on the gradient signals. To illustrate the advantage, consider a scenario where the EMA teacher generates a false positive (FP) sample, while the DSOD teacher correctly rejects it. A naive box-level fusion would retain this FP in the final pseudo-label, and the MT could not fix this error. In contrast, our loss-level fusion dynamically down-weights the contribution of such FP samples by leveraging the corrective signal from the DSOD teacher, thereby mitigating error propagation.

Specifically, we introduce three loss functions  $\mathcal{L}_{\text{hard}}^{\text{EMA}}$ ,$\mathcal{L}_{\text{hard}}^{\text{EMA}}$  and $\mathcal{L}_{\text{soft}}^{\text{Static}}$   to learn from the hard labels provided by the EMA teacher, the hard labels from the static teacher, and the soft labels from the static teacher, respectively. The former two are normal detection loss and the last $\mathcal{L}_{\text{soft}}^{\text{Static}}$  is motivated by logit distillation~\cite{detrdistill}:
\begin{equation}
    \label{eq:soft_static}
     \mathcal{L}_{\text{soft}}^{\text{Static}} = \sum_i \Bigl[ \alpha \cdot \mathcal{L}_{\text{bce}}(c^T_{\sigma(i)} , c^S_{i}) + \beta \cdot \mathcal{L}_{\text{bbox}}(b^T_{\sigma(i)},b^S_{i}) \Bigl],
\end{equation}
where $\sigma(i)$ denotes the optimal assignment between teacher and student predictions, $c_i$ and $b_i$ represent the classification scores and bounding‑box coordinates of the $ith$ prediction, respectively, $\mathcal{L}_{\text{bce}}$  represent the binary cross-entropy loss,  $\mathcal{L}_{\text{bbox}}$ is a combination of L1 loss and GIoU loss, and $\alpha$, $\beta$ are balancing coefficients. The superscripts $T$ and $S$ denote the teacher and student models, respectively. 
We set $\alpha=5$ and $\beta=2$ to achieve effective optimization.

Furthermore, we observe that initializing both the student and teacher models with weights from the DSOD model yields a significant performance gain (Sec.~\ref{sec:ablation}). This is because the VFM‑independent parameters in the DSOD model have already undergone partial adaptation to the target domain, thereby providing a better starting point for the distillation process.

\subsection {Training Objective}

In the setting of VFM-assisted adaptation, we follow previous work DRU~\cite{dru} to use Masked Image Consistency (MIC)~\cite{mic} loss to facilitate the learning of context information and better facilitate the fusion of dual encoders. The optimization target is given as follows:
\begin{equation}
    {L}_{\text{adapt}} = {\mathcal{L}}_{\text{det}}(\text{pred}, \text{PL}) + {\mathcal{L}}_{\text{det}}(\text{pred}^M, \text{PL})  + \lambda_1 \cdot {\mathcal{L}}_{\text{reg}},
\end{equation}
where $\text{pred}$, $\text{pred}^{M}$ represents the predictions of the normal branch and the masked branch, respectively and $\lambda_1$ is a balancing factor.  $\text{PL}$ denotes the pseudo-labels. ${\mathcal{L}}_{\text{det}}$ represents a standard DETR~\cite{detr,deformabledetr} detection loss, as defined in Sec.~\ref{sec:detr}.

In the two‑stage distillation setting, the loss function is defined as:
\begin{equation}
    {L}_{\text{distill}} = \mathcal{L}_{\text{hard}}^{\text{EMA}} + \lambda_2 \cdot \mathcal{L}_{\text{hard}}^{\text{Static}} + \lambda_3 \cdot \mathcal{L}_{\text{soft}}^{\text{Static}},
\end{equation} 
where $\mathcal{L}_{\text{hard}}^{\text{EMA}}$ and ${L}_{\text{hard}}^{\text{Static}}$ are the adaptation loss $ \mathcal{L}_{\text{adapt}} $ computed with pseudo-labels from EMA teacher static teacher, respectively,
$\lambda_2$ and $\lambda_3$ are two balancing weights that control the contribution of static teacher.

\section{Experiments}
We conduct comprehensive experiments to evaluate the effectiveness of our method on multiple standard SFOD benchmarks. Then, we perform ablation studies by using different exchange strategies to validate the effectiveness of the proposed method. 
Finally, we analyze the promising results of our method through detailed visualization and component analysis.

\subsection{Datasets and Task settings}
We use four public datasets in the following experiments: 1) \emph{Cityscapes}~\cite{cityscape} is a classical urban dataset that  comprising 5,000 images with high-quality pixel level annotations. We use 2,925 training images and 500 validation images in the following experiments. 2) \emph{Foggy Cityscapes}~\cite{foggycityscape} is derived from \emph{Cityscapes}, this dataset incorporates synthetic fog simulation across three density levels (0.005, 0.01, 0.02) to simulate adverse weather conditions. We use the strongest level 0.02 for the experiments. 3) \emph{BDD100k}~\cite{bdd100k} is a large-scale driving dataset containing 100,000 images captured under diverse weather conditions, times of day, and driving scenarios. 4) \emph{SIM10k}~\cite{sim10k} is a synthetic dataset consisting of 10,000 computer-generated urban scenes with detailed car annotations.

Following previous works~\cite{dru,a2sfod,pets}, we validate our method on three representative SFOD tasks including three types of domain shift:
\begin{enumerate}
\item \textbf{Normal to Foggy Adaptation}: Adaptation from clear weather conditions to foggy environments, using \emph{Cityscapes} as the source domain and \emph{Foggy Cityscapes} as the target domain.
\item \textbf{Cross Scene Adaptation}: Generalization across different real-world driving scenarios, with \emph{Cityscapes} as the source domain and \emph{BDD100k} as the target domain.
\item \textbf{Synthetic to Real Adaptation}: Transfer from synthetic imagery to real-world data, employing \emph{SIM10k} as the source domain and \emph{Cityscapes} as the target domain.
\end{enumerate}
\begin{table*}[htbp]
  \centering
  \resizebox{0.85\textwidth}{!}{
  \begin{tabular}{@{}c|c|c|cccccccc|c@{}}
    \toprule
    &Method & Detector & person & rider & car & truck & bus & train & motor & bicycle & mAP\\
    \midrule
    \midrule
    &Source Only & Def.DETR & 40.0 & 41.2 & 47.0 & 13.0 & 29.1 & 6.5 & 21.5 & 38.0 & 29.5 \\
    \midrule
    \multirow{8}{*}{\rotatebox[origin=c]{90}{UDAOD}}
    &SW-Faster \cite{swfaster} & FRCNN &  32.3 & 42.2 & 47.3 & 23.7 & 41.3 & 27.8 & 28.3 & 35.4 & 34.8 \\
    &CR-DA-DET \cite{crdadet} & FRCNN & 32.9 & 43.8 & 49.2 & 27.2 & 45.1 & 36.4 & 30.3 & 34.6 & 37.4 \\
    &TIA \cite{tia} & FRCNN & 34.8 & 46.3 & 49.7 & 31.1 & {52.1} & {48.6} & { 37.7} & 38.1 & 42.3 \\
    &PT \cite{pt} & FRCNN & 40.2 & 48.8 & 59.7 & 30.7 & 51.8 & 30.6 & 35.4 & 44.5 & 42.7 \\
    &TDD \cite{tdd} & FRCNN & 39.6 & 47.5 & 55.7 & {33.8} & 47.6 & 42.1 & 37.0 & 41.4 & 43.1 \\
 &RobustDet~\cite{robust_object_detection} & FRCNN & 39.9 &47.5 & 59.2 & 31.9 & 52.0 & 37.8 & {38.4} & 38.6 &43.2 \\ 
 &DSCA~\cite{dsca} & FRCNN & \underline{53.9} & \underline{61.3} & \underline{68.8} & 32.5 & 52.6 & 24.3 & 35.2 & 52.7 & {47.6} \\
    &MIC~\cite{mic} & FRCNN & 50.9 & 55.3 & 67.0 & 33.9 & 52.4 & 33.7 & 40.6 &47.5 & 47.6\\
    &AT~\cite{at} & FRCNN & 45.5 & 55.1 & 64.2 & 35.0 & 56.3 & \underline{54.3} & 38.5 & 51.9 & 50.9\\
    &DT~\cite{dinoteacher} & FRCNN & {48.5} & {60.0} & {65.4} & \underline{47.2} & \underline{66.5} & {52.9} & \underline{46.2} & \underline{56.7} & \underline{55.4} \\
    &SFA \cite{sfa} & Def.DETR & 46.5 & 48.6 & 62.6 & 25.1 & 46.2 & 29.4 & 28.3 & 44.0 & 41.3 \\
    &MTTrans \cite{mttrans} & Def.DETR & 47.7 & 49.9 & {65.2} & 25.8 & 45.9 & 33.8 & 32.6 & {46.5} & 43.4 \\
    &DA-DETR \cite{dadetr} & Def.DETR &  {49.9} & {50.0} & 63.1 & 24.0 & 45.8 & 37.5 & 31.6 & 46.3 & {43.5} \\
    &$\text{O}^2$net~\cite{o2net} & Def.DETR & 48.7 & 51.5 & 63.6 & 31.1 & 47.6 &47.& 38.0 & 45.9 & 46.8\\
   
    \midrule
    \multirow{6}{*}{\rotatebox[origin=c]{90}{SFOD}}&SED(Mosaic) \cite{sed} & FRCNN & 33.2 & 40.7 & 44.5 & 25.5 & 39.0 & 22.2 & 28.4 & 34.1 & 33.5\\
    &A${^2}$SFOD \cite{a2sfod} & FRCNN & 32.3 & 44.1 & 44.6 & 28.1 & 34.3 & 29.0 & 31.8 & 38.9 & 35.4 \\
    &LODS \cite{lods} & FRCNN & 34.0 & 45.7 & 48.8 & 27.3 & 39.7 & 19.6 & 33.2 & 37.8 & 35.8 \\
    &PETS \cite{pets} & FRCNN & 42.0 & 48.7 & 56.3 & 19.3 & 39.3 & 5.5 & 34.2 & 41.6 & 35.9 \\
    &IRG \cite{irg} & FRCNN & 37.4 & 45.2 & 51.9 & 24.4 & 39.6 & 25.2 & 31.5 & 41.6 & 37.1 \\
    &LPLD \cite{lpld} & FRCNN & 39.7 & 49.1 & 56.6 & 29.6 & 46.3 & 26.4 & 36.1 & 43.6 & 40.9 \\ 
    &DRU \cite{dru} & Def.DETR & 48.3 &  51.5 & 62.5 & 26.2 & 43.2 & 34.1 & 34.2 & 48.6 & 43.6 \\
    &DDT \cite{ddt} & Def.DETR  & 49.3 & {\bf 53.0} & 65.4 & 25.8 & 43.0 & {\bf 39.7} & {\bf 40.0} &47.0 &45.5 \\
    \rowcolor{gray!19}
    \cellcolor{white}&DSOD-Distill  & Def.DETR  & {\bf 52.1} & 51.1 & {\bf 67.9} & 32.6 &45.1 &  38.5 & 35.1 & {\bf 49.6} & 46.9\\
    \rowcolor{gray!19}
    \cellcolor{white}&DSOD & Def.DETR & 48.5 & 48.1 & 64.6 & {\bf 35.8} & {\bf 51.1} & 38 &  36.6 & 48.3 & {\bf 48.1} \\
    \midrule
    &Oracle & Def.DETR & 52.9 & 52.9 & 71.8 &  31.3 & 53.9 & 41.0 & 41.4 & 44.0 & 48.7 \\

  \bottomrule
  \end{tabular}
  }
  \caption{Comparison results (\%) of different UDAOD and SFOD methods under the setting of Normal to Foggy Adaptation (Cityscapes $\to$ Foggy Cityscapes). ``Source Only and  ``Oracle'' refers to the source-trained and target-trained model, respectively. FRCNN denotes Faster R-CNN~\cite{faster}, and Def.DETR represents Deformable DETR~\cite{deformabledetr}. } 
  \label{tab:city2foggy}
  
\end{table*}

\subsection{Implementation Details}
Following established practices~\cite{dru}, we employ Deformable-DETR~\cite{deformabledetr} with a ResNet-50 backbone~\cite{he2016deep} pre-trained on ImageNet~\cite{imagenet_cvpr09} as our base detection architecture. For feature extraction, we employ DINOv2~\cite{oquab2023dinov2} with a ViT-B/Reg backbone across all experiments.  In line with DRU~\cite{dru}, we incorporate strategies such as dynamic retraining and masked image consistency, while keeping the same hyperparameter configurations. We use  a batch size of 8 and initialize our model with the source-pretrained weights provided by DRU. 

For input preprocessing, we adopt distinct resizing strategies for the two backbones. For the CNN backbone, images are resized with the shorter edge scaled to 600 pixels while preserving the original aspect ratio. For the DINOv2 backbone, images are further resized to half the dimensions of the CNN-processed inputs, effectively balancing computational efficiency and feature quality. A larger input size could yield better results but also increase the training burden.
We apply a comprehensive set of data augmentation techniques including random erasing, random horizontal flipping, and color transformations to enhance model robustness and generalization capability following previous works~\cite{irg,dru,pets}.

In the first training stage, the model is trained for 30 epochs for Normal-to-Foggy adaptation and Synthetic-to-Real adaptation, and for 5 epochs for Cross-Scene Adaptation. In the distillation stage, all models are trained for 5 epochs. The best results are reported.
In our empirical evaluation, we set $\lambda_1 =0.1, \lambda_2=0.5, \lambda_3=0.5$ to achieve the best results.
Model effectiveness is quantified using the mean average precision (mAP) metric with an IoU threshold of 0.5, following standard evaluation protocols in DAOD literature~\cite{irg,dru,dadetr,dafaster}.

\begin{table*}[htbp]
  \centering
  \resizebox{0.85\textwidth}{!}{
  \begin{tabular}{@{}c|c|c|c c c c c c c|c@{}}
    \toprule
    &Method & Detector & truck & car & rider & person & motor & bicycle & bus & mAP \\
    \midrule
    \midrule
    &Source Only & Def.DETR & 18.9 & 58.2 & 28.3 & 42.0 & 15.7 & 18.8 & 21.7 & 29.1 \\
    \midrule
    \multirow{7}{*}{\rotatebox[origin=c]{90}{UDAOD}}
    &DA-Faster \cite{dafaster} & FRCNN & 14.3 & 44.6 & 26.5 & 29.4 & 15.8 & 20.6 & 16.8 & 24.0 \\
    &SW-Faster \cite{swfaster}& FRCNN & 15.2 & 45.7 & 29.5 & 30.2 & 17.1 & 21.2 & 18.4 & 25.3 \\
    &CR-DA-DET \cite{crdadet} & FRCNN & 19.5 & 46.3 & 31.3 & 31.4 & 17.3 & 23.8 & 18.9 & 26.9 \\
    &PT~\cite{pt} & FRCNN & 25.8 & 52.7 & 39.9 & 40.5 & 23.0 &28.8 & 33.8 & 34.9 \\
    &DT~\cite{dinoteacher} & FRCNN & \underline{44.3} & \underline{66.6} & \underline{47.0} & \underline{51.6} & \underline{38.3} & \underline{40.8} & \underline{45.9}  & \underline{47.8} \\
    &AQT \cite{aqt} & Def.DETR & 17.3 & 58.4 &{33.0} & 38.2 & 16.9 & 23.5 & 18.4 & 29.4 \\
    &O$^2$net \cite{o2net} & Def.DETR & 20.4 & 58.6 & 31.2 & 40.4 & 14.9 & 22.7 & 25.0 & 30.5 \\
    &MTTrans \cite{mttrans} & Def.DETR & {25.1} & 61.5 & 30.1 & 44.1 & 17.7 & {23.0} & {26.9} & 32.6 \\
    &MRT \cite{mrt} & Def.DETR & 24.7 & {63.7} & 30.9 &{ 48.4} & {20.2} & 22.6 & 25.5 & {33.7} \\
    \midrule
    \multirow{4}{*}{\rotatebox[origin=c]{90}{SFOD}}
    &SED(Mosaic) \citep{sed} & FRCNN & 20.6 & 50.4 & 32.6 & 32.4 & 18.9 & 25.0 & 23.4 & 29.0 \\
    &PETS \protect\cite{pets} & FRCNN & 19.3 & 62.4 & 34.5 & 42.6 & 17.0 & 26.3 & 16.9 & 31.3 \\
    &A${^2}$SFOD \cite{a2sfod} & FRCNN & 26.6 & 50.2 & 36.3 & 33.2 & 22.5 & 28.2 & 24.4 & 31.6 \\
    &DRU~\cite{dru} & Def.DETR & 27.1 & 62.7 &  36.9 & 45.8 & 22.7 & 32.5 &  28.1 & 36.6 \\
    &DDT~\cite{ddt} & Def.DETR  & \textbf{28.7} & \textbf{66.1} & \textbf{40.3} & \textbf{49.3} & \textbf{29.3} &34.1 &  29.1 & {\bf 39.6} \\
   \rowcolor{gray!19}
    \cellcolor{white}&DSOD-Distill & Def.DETR &  25.7 & 63.9 & 35.7 & 47.5 & 23.3 & 32.9 & 25.0 &  36.7 \\
   \rowcolor{gray!19} 
    \cellcolor{white}&DSOD & Def.DETR &  28.5 & 61.6 &  36.4 & 47.8 &  28.0 & \bf{ 34.9} & {\bf 32.2} &  39.2 \\
    \midrule
    & Oracle &  Def.DETR & 66.9 & 87.9 & 56.4 & 74.9 & 53.8 & 68.3 & 55.0 & 66.2\\
  \bottomrule
  \end{tabular}
  }
  \caption{Comparison results (\%) of Cross Scene Adaptation (Cityscapes $\to$  BDD100k)}
  \label{tab:city2bdd}
  
\end{table*}

\begin{table}[htb]
  \centering
     \centering
     \scalebox{0.7}{
     \begin{tabular}{@{}c|c|c|c@{}}
         \toprule
         &Method & Detector & AP of car \\
         \midrule
         \midrule
         &Source Only & Def.DETR & 48.9 \\
         \midrule
        \multirow{7}{*}{\rotatebox[origin=c]{90}{UDAOD}}
         &DSCA~\cite{dsca} & FRCNN & 50.0 \\
         &TDD \cite{tdd} & FRCNN & 53.4 \\
         &PT \cite{pt} & FRCNN & 55.1 \\
         &SFA \cite{sfa} & Def.DETR & 52.6 \\
         &O$^2$net \cite{o2net} & Def.DETR & 54.1 \\
         &DA-DETR \cite{dadetr} & Def.DETR & 54.7 \\
         &MTTrans \cite{mttrans} & Def.DETR & 57.9 \\
         &MTM \cite{mtm} & Def.DETR & \underline{58.1} \\
         \midrule
         \multirow{5}{*}{\rotatebox[origin=c]{90}{SFOD}}
         &SED(Mosaic) \cite{sed} & FRCNN & 43.1 \\ 
         &IRG \cite{irg} & FRCNN & 43.2 \\ 
         &A${^2}$SFOD \cite{a2sfod} & FRCNN & 44.0 \\
         &LPU~\cite{lpu} & FRCNN & 47.3 \\
         &LPLD~\cite{lpld} & FRCNN & 49.4 \\
         &PETS \cite{pets} & FRCNN & 57.8 \\
         
         &DRU~\cite{dru} & Def.DETR & 58.7 \\
         &DDT~\cite{ddt} & Def.DETR & 60.6 \\
         \rowcolor{gray!19}
         \cellcolor{white}&DSOD-Distill & Def.DETR & 60.6 \\
        \rowcolor{gray!19}
         \cellcolor{white} &DSOD & Def.DETR & \textbf{61.4} \\
         \midrule
         & Oracle & Def.DETR  & 75.9 \\
        \bottomrule
     \end{tabular}
     }
     \caption{Comparison results (\%) of Synthetic to Real Adaptation (Sim10k $\to$  Cityscapes)}
     \label{tab:synthetic}
  
\end{table}
\subsection{Comparison with SOTA Methods}
UDAOD and SFOD share a similar task formulation. Therefore, we compare our method with existing UDAOD and SFOD approaches. The results are summarized in Tables~5.1-5.3, where “Source only” and “Oracle” represent models that are trained exclusively on source domain or target domain data, respectively. These two baselines establish the lower and upper performance bounds for the SFOD task.

\noindent\textbf{Normal to Foggy Adaptation}.
We benchmark our method against state-of-the-art (SOTA) UDAOD and SFOD approaches, and demonstrate that it outperforms most existing methods. As shown in Table~\ref{tab:city2foggy}, our DSOD surpasses UDAOD methods DA-DETR~\cite{dadetr} by a notable margin of 4.5\% AP and $\text{O}^2$net~\cite{o2net} by a 1.2\% AP. As for the  SFOD methods, our methods leads the baseline DRU~\cite{dru} by 4.5\% AP and the very recent competitor DDT~\cite{ddt} by 2.6\% AP. These results show that by harnessing the power of VFMs, source-free methods can achieve highly competitive performance. 
Moreover, we observe that with an appropriate distillation strategy, our DSOD-Distill variant remains highly effective, with only a marginal drop of 1.1\% AP in performance and still surpasses DDT by 1.4\% AP.
It is noteworthy that DSOD-Distill even performs better than the original DSOD in some categories such as ``person'' and ``car'', which validates the potential of such two-stage method in generating high-quality pseudo-labels. DSOD achieves state-of-the-art performance.

\noindent\textbf{Cross Scene Adaptation}.
In cross-scene settings, our method continues to demonstrate strong performance, as detailed in Table~\ref{tab:city2bdd}. Our approach outperforms the UDAOD method MRT~\cite{mrt} by 4.9\% AP and surpasses DRU by 2\% AP, underscoring its effectiveness in cross-scene adaptation. Even when compared with the SOTA method DDT~\cite{ddt}, our method still exhibits competitive advantages, performing notably better in several categories such as ``bicycle'' (leading by 0.8\% AP) and ``bus'' (leading by 3.1\% AP). Although not achieving overall SOTA performance, our method possesses two significant advantages: 1) Scalability: it can readily benefit from larger VFM backbones and higher input resolutions; and 2) Compatibility: it has strong potential for seamless integration and joint optimization with existing approaches like DDT.

\noindent\textbf{Synthetic to Real Adaptation}.
The results on synthetic-to-real adaptation are presented in Table~\ref{tab:synthetic}, where our method achieves SOTA performance. Specifically, our methods surpass all of UDAOD methods including PT~\cite{pt} (+6.3\% AP), MTTrans~\cite{mttrans} (+3.5\% AP) and MTM~\cite{mtm} (+3.3\% AP). Then, compared to the strong SFOD competitor DDT~\cite{ddt}, our approach attains an improvement of 0.8\% AP. Moreover, even the VFM-free variant, DSOD-Distill, demonstrates performance comparable to DDT. These findings further confirm the efficacy of our method in synthetic-to-real domain adaptation.

\subsection{Ablation Studies}\label{sec:ablation}
\begin{table}[htb]
    \centering
    \scalebox{0.8}{
        \begin{tabular}{ccccc|c}
    \toprule
        MT & UFI &  SAFR & DDAW & DRU & AP \\
    \midrule
         &  &  & & & 29.5 \\ 
    \midrule
         \checkmark & & & &  &39.6 \\ 
         \checkmark & \checkmark & & &  &43.0 \\ 
         \checkmark & \checkmark & \checkmark & & & 45.5 \\ 
         \checkmark &  \checkmark   & \checkmark & \checkmark &  &45.8 \\ 
         \rowcolor{gray!19}
         \checkmark & \checkmark    & \checkmark & \checkmark & \checkmark & \textbf{ 48.1} \\ 

    \bottomrule
    \end{tabular}}
    

    
    \caption{Ablation study (\%) of the proposed component UFI, SAFR, DDAW in the setting of cross scene adaptation. ``MT'' refers to the Mean Teacher baseline and ``DRU'' refers to the setting that uses proposed method in DRU. We use our reproduced ``MT'' results for fair comparison. }
    \label{tab:abl_component}
\end{table}
\noindent\textbf{Effects of each component.}  To validate the effectiveness of the proposed components, we conduct comprehensive ablation studies on the Normal to Foggy adaptation setting (Cityscapes $\to$ Foggy Cityscapes). The results are summarized in Table~\ref{tab:abl_component}. The first row is the source-pretrained model and the second row is our reproduced Mean Teacher baseline. Our proposed UFI module contributes 3.4\% AP over the baseline. This demonstrates that a simple scale extension and addition fusion already can help the pre-trained detector benefit from the DINOv2's representative features by preserving most of the original information. The proposed SAFR module  further brings 2.5\% AP gain, validating the effectiveness of semantic-aware feature regularization. The DDAW strategy improves 0.3\% AP which seems to be marginal but without which, the hyper-parameter selection would cost much time. Finally, combined with DRU, our method is further improved by 2.3\% AP, confirming that our improvements are orthogonal to DRU.  These results validate the effectiveness of our proposed methods.  


\begin{table}[htb]
    \centering
    \scalebox{0.7}{
    \begin{tabular}{c|ccccccc}
    \toprule
        Projection & MLP  &  De-convolution & Sp.Attn & Ch.Attn & Linear-s & Learnable SSE &  SSE\\
        \midrule
        AP & 42.3 & 40.5 & 41.7 & 42.4 & 41.4 & 40.1 & \cellcolor{gray!19}\textbf{43.2}  \\
        \bottomrule
        
    \end{tabular}
    }
    \caption{Ablation study results (\%) of the projection module in the setting of normal to foggy adaptation. ``Linear-s'' refers to a single-scale linear projection module and other modules are all implemented in multi-scale setting. ``De-convolution'' is implemented as ViT-Det suggested. ``Sp.Attn'' and ``Ch.Attn'' refer to the spatial attention and channel attention.}
    \label{tab:abl_proj}
\end{table}

\noindent \textbf{Ablation of projection module.} We conduct an ablation study on the design of the projection module, with results shown in Table~\ref{tab:abl_proj}. We implement several projection variants, including an MLP block, a de‑convolution block, and two types of attention‑based block, \emph{i.e.} spatial attention block and channel attention block. The proposed SSE module implemented through linear layer achieves the best performance (43.2\% AP), indicating that more complex modules face challenges in optimization and are prone to catastrophic forgetting on the limited target‑domain data. Additionally, we implement a single‑scale variant of SSE, denoted as ``Linear‑s", which performs 1.8\% AP lower than the multi‑scale version. We also implement a Learnable SSE that has learnable fusion weight, which performs 3.1\% AP lower, indicating that the model is affected by overfitting.  This result further confirms the effectiveness of multi‑scale feature fusion in aligning ResNet~\cite{he2016deep} and DINOv2~\cite{oquab2023dinov2} representations.

\begin{figure}
    \centering
    \includegraphics[width=0.8\linewidth]{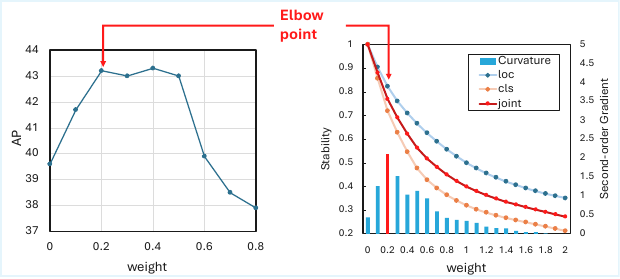}
    \caption{\textbf{Left}: The AP (\%) variation with fusion weight $w$ in the setting of Normal to Foggy adaptation.  The \textcolor{red}{red arrow} indicates the elbow point selected based on the curvature analysis. \textbf{Right}: The stability variation with fusion weight $w$. ``loc'' represents localization stability, ``cls'' represents classification stability and ``joint'' represents joint stability. The \textcolor{red}{red bar} marks the point of maximum curvature (elbow point).}
    \label{fig:fig_ap_curve}
\end{figure}
\noindent \textbf{The effect of DDAW}. We visualize the weight selection process for the Normal-to-Foggy adaptation scenario in Fig.~\ref{fig:fig_ap_curve}. The comparison between the stability curve and AP  curve validates our selection strategy, showing strong alignment between model performance and stability metrics. We observe that the classification score exhibits greater sensitivity than the localization score, as evidenced by its sharper decline in the stability curve. Furthermore, the sharp decline in the AP curve beyond the weight threshold of 0.4 indicates that training instability becomes the dominant factor, catastrophically degrading model performance. This suggests that excessive incorporation of VFM features beyond this critical point disrupts the feature manifold, overwhelming the model's capacity for stable adaptation.

\noindent \textbf{Ablation of distillation process.}
We conduct an ablation study of the distillation process, with results summarized in Table~\ref{tab:abl_distill}. As shown, the hard label loss from the static teacher, $\mathcal{L}_{\text{hard}}^{\text{EMA}}$ , contributes the most, improving performance by 4.9\% AP over the MT baseline where only $\mathcal{L}_{\text{hard}}^{\text{EMA}}$ is enabled. The soft loss from the static teacher and DSOD initialization further contribute 1.6\% AP and 0.8\% AP, respectively. It proves that not only the final results but also the intermediate predictions of the DSOD teacher is helpful to the distillation. Also, our proposed method surpasses the box-level fusion baseline largely with 2.6\% AP. These results validate the effectiveness of the proposed distillation framework.
\begin{table}[htbp]
    \centering
    \scalebox{1.0}{
    \begin{tabular}{cccc|c}
    \toprule
        $\mathcal{L}_{\text{hard}}^{\text{EMA}}$ & $\mathcal{L}_{\text{hard}}^{\text{Static}}$  & $\mathcal{L}_{\text{soft}}^{\text{Static}}$ & DSOD Init  & AP \\
    \midrule
         &  &  & & 29.6 \\ 
    \midrule
         \checkmark & & & & 39.6 \\ 
         \checkmark & \checkmark & & & 44.5 \\ 
         \checkmark & \checkmark & \checkmark &  & 46.1 \\
         \checkmark & \checkmark &  & \checkmark & 46.0 \\ 
         \rowcolor{gray!19}
         \checkmark & \checkmark & \checkmark & \checkmark & 46.9 \\ 
    \midrule
    \multicolumn{3}{l}{Box fusion}   & & 44.3 \\
    \multicolumn{3}{l}{DSOD Teacher}   & & 48.1\\
    \bottomrule
    \end{tabular}}
    \caption{Ablation study results (\%) of the distillation process in the setting of normal to foggy adaptation. ``DSOD Init'' refers to using the DSOD weights for initialization. ``Box fusion'' represents using NMS for box-level fusion baseline.  }
    \label{tab:abl_distill}
\end{table}

\subsection{Analysis and Discussion}\label{sec:visual_analysis}

\begin{figure}[th]
    \centering
    \includegraphics[width=1\linewidth]{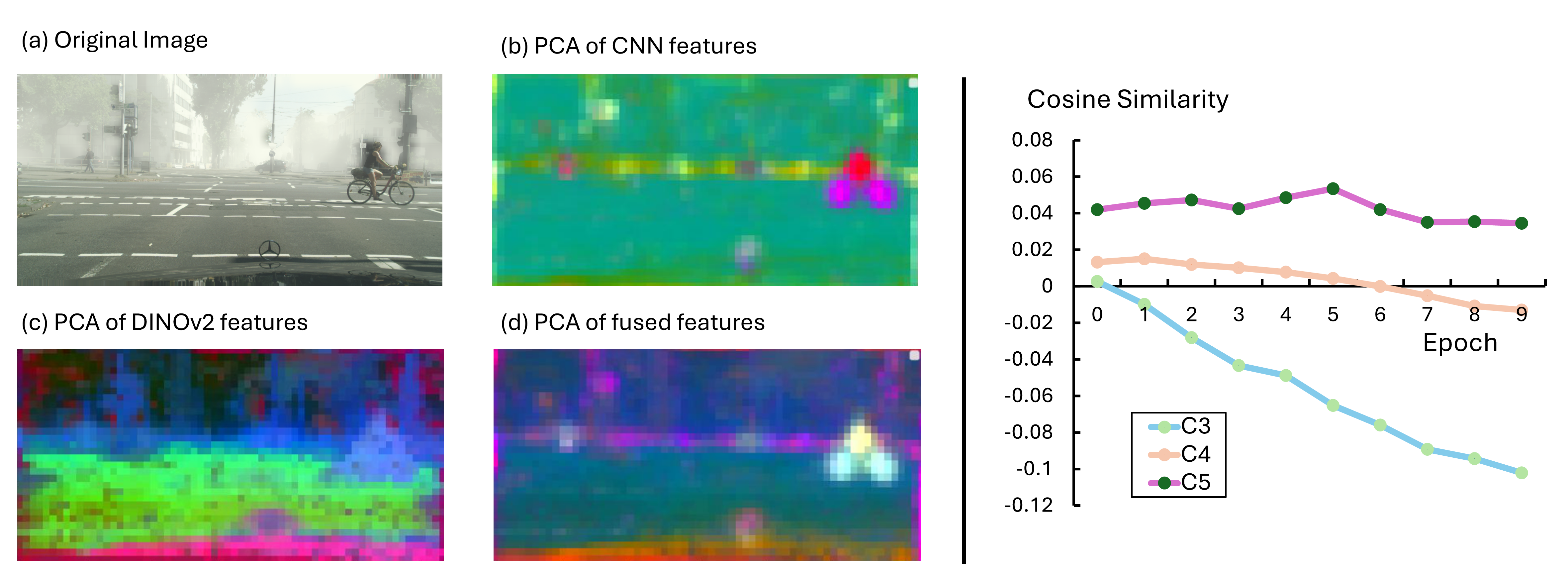}
    \caption{\textbf{Left:} PCA visualization and comparison of features. Color intensity represents feature activation magnitude.
    \textbf{Right:} Change of cosine similarity  between CNN features and projected DINOv2 features over the model training process.  }
    \label{fig:orthogonal_vis}
\end{figure}

\begin{figure}[htbp]
    \centering
    \includegraphics[width=1.0\linewidth]{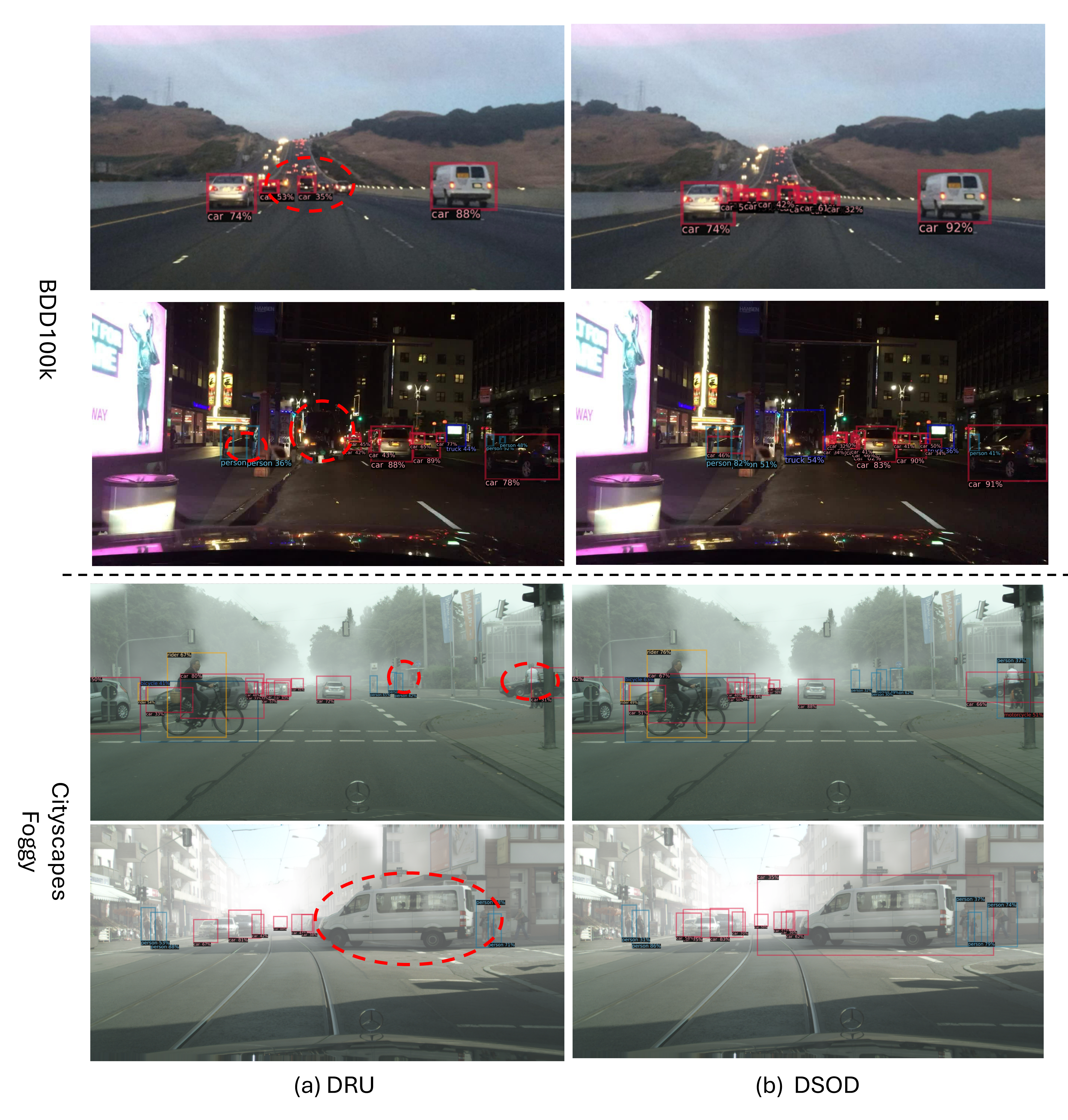}
    \caption{Qualitative comparison of detection performance. Subfigures (a) and (b) present the visualization results of DRU~\cite{dru} and our proposed DSOD, respectively. It can be observed that DSOD achieves more complete detection, as evidenced within the \textcolor{red}{\emph{red dashed circles}} where DRU fails to identify certain objects. }
    \label{fig:visualize_pred}
\end{figure}

\noindent \textbf{The orthogonality of features}. Random high-dimensional vectors tend to be orthogonal~\cite{vershynin2018high}. However, this property may to hold for VFM features and CNN features since they are not i.i.d. To validate this point, we perform  two analysis experiments. \textbf{First}, we conduct the Principle Component Analysis (PCA) decomposition on the feature representations: $\mathbf{F}_{\text{CNN}}$, $\mathbf{F}_{\text{DINO}}^{proj}$, and $\mathbf{F}_{\text{fuse}}$. As visualized  in Fig.~\ref{fig:orthogonal_vis}, $\mathbf{F}_{\text{CNN}}$ exhibits tight instance-level clustering, while   $\mathbf{F}_{\text{DINO}}^{proj}$  shows semantic-level grouping. Crucially, $\mathbf{F}_{\text{fuse}}$ retains both structural patterns, confirming effective information preservation through our simple but effective fusion strategy.
\textbf{Second}, we quantify this orthogonal relationship by computing the cosine similarity between the CNN features $\mathbf{F}_{\text{CNN}}$ and the projected DINOv2 features $\mathbf{F}_{\text{DINO}}^{proj}$ across different levels ($C_3$, $C_4$, $C_5$), finding that they maintain an almost orthogonal relationship (cosine similarity < 0.1). Notably, $C_3$ curve exhibits a decreasing trend and deviates from zero, which can be attributed to the significant semantic gap between low-level features and VFM features. 
These findings highlight that, \emph{through a simple addition operation, we can effectively preserve knowledge from both the CNN and the VFM}.

\noindent \textbf{Qualitative comparison}. Fig.~\ref{fig:visualize_pred} provides a qualitative comparison of detection results. Our DSOD consistently outperforms the baseline (DRU) in recall, handling diverse object categories effectively across various environmental conditions. The red circles highlight regions where DSOD successfully detects significant objects missed by DRU. We attribute this performance gap to the limited robustness of the baseline in challenging scenarios. In contrast, DSOD benefits from the semantic complementarity and unbiased feature representations provided by VFMs, making it more resilient to such difficult conditions.

\section{Conclusion}

In this paper, we propose DSOD, a novel framework that leverages VFMs to achieve robust and unbiased SFOD. DSOD comprises two key components: (1) the Unbiased Feature Injection  module, which injects robust VFM features into the detection model, and (2) the Semantic-Aware Feature Regularization  module, which enhances the regularization effect of VFMs. To stabilize mean-teacher training, we introduce a Domain-Aware Adaptive Weighting  strategy that intelligently adjusts fusion weights throughout the learning process. Furthermore, we devise a Dual-Teacher distillation scheme to transfer knowledge into a lightweight, VFM-free variant, DSOD-Distill, enabling resource-efficient deployment. Extensive experiments on multiple benchmarks validate the effectiveness of the proposed approach and its components.



\bibliographystyle{elsarticle-num} 
\bibliography{elsarticle}
\end{document}